# A novel segmentation dataset for signatures on bank checks

Muhammad Saif Ullah Khan
*Department of Computer Science*
*University of Kaiserslautern*
Kaiserslautern, Germany
khanm@rhrk.uni-kl.de

*Abstract*—The dataset presented provides high-resolution images of real, filled out bank checks containing various complex backgrounds, and handwritten text and signatures in the respective fields, along with both pixel-level and patch-level segmentation masks for the signatures on the checks. The images of bank checks were obtained from different sources, including other publicly available check datasets, publicly available images on the internet, as well as scans and images of real checks. Using the GIMP graphics software, pixel-level segmentation masks for signatures on these checks were manually generated as binary images. An automated script was then used to generate patch-level masks. The dataset was created to train and test networks for extracting signatures from bank checks and other similar documents with very complex backgrounds.

I. SPECIFICATION TABLE

| | |
|---|---|
| Subject | Computer Vision and Pattern Recognition |
| Specific subject area | Handwritten signature segmentation from documents |
| Type of data | Image |
| How data were acquired | The dataset consists of images of 158 bank checks, which were acquired from different sources. The first 37 images, which consist of either counterfeit checks used in various scams or other sample bank checks, were downloaded from the Internet via Google Image Search. Next 9 images are scanned, real or dummy checks. The remaining 110 checks come from a publicly available dataset [2]. These images were then used to manually generate the segmentation data. |
| Data format | Raw |
| Parameters for data collection | Images which showed a complete bank check and contained at least one handwritten signature in the signature field were included in the dataset. Most of the checks also contain other handwritten text besides the signature. |
| Description of data collection | After acquiring the images of bank checks, all images were resized such that the longer side of each image is 2240 pixels with a resolution of 300 pixels/inch. Labels in form of pixel-level segmentation masks of signatures for all the images were manually generated using an image processing tool. The segmentation mask is a black-and-white image with the pixels containing the signature shown in white and the rest of the check shown in black. A Python script was then used to generate patch-level segmentation masks by estimating bounding boxes around the signature pixels. |
| Data source location | Institution: University of Kaiserslautern<br>City: Kaiserslautern<br>Country: Germany<br><br>Primary data sources:<br>The original images which were used to create the dataset come from multiple sources, with a major primary source being the IDRBT Cheque Image Dataset [2]. |
| Data accessibility | Repository name: Kaggle repository [1]<br>Data identification number: 10.34740/KAGGLE/DS/1041694<br>Direct URL to data: https://doi.org/10.34740/KAGGLE/DS/1041694 |

II. VALUE OF DATA

Despite extensive research into automatic signature verification, end-to-end signature verification systems in banks are not widely available yet. This is primarily because of a lack of an efficient method of extracting signatures from bank checks before feeding them to a verification system. This dataset is important in this regard, as there are not many publicly available datasets which deal with extracting signatures specifically from bank checks.

The data is useful for the researchers working on segmentation of bank checks or other similar documents with highly complex backgrounds, to extract handwritten signatures from them.

This data might be used to train different neural networks for completely automatic and accurate, pixel-level segmentation of signatures from bank check images or any other document, which is an important component of any end-to-end signature verification system.

## III. DATA DESCRIPTION

The dataset is divided into a training set of 129 images and a test set of 27 images. In each of these subsets, there is a folder X containing the RGB color images of bank checks. And another folder y which has two subfolders, called pixel and patch, containing the pixel-level and patch-level segmentation masks respectively for the corresponding check image in the X folder. Each segmentation mask is a black-and-white image with the same dimensions as the corresponding check in the dataset. Signature pixels (and patches respectively) are labelled with white color, whereas the rest of the check including background, logos, printed text and any other handwritten text are all labelled with black color. For patch segments, a minimal bounding box is drawn around the signature pixels, and then the whole box is colored white. Most of the checks have one signature each, but there are a few with more than one signature per image. For such checks, if the bounding boxes around the individual signatures overlap, they are labeled as a single patch. For disjoint bounding boxes, there are multiple labeled patches.

All images are in JPEG format and were resized such that the longer size of the image is exactly 2240 pixels long. The images in the test set were all downloaded from Google Image Search and originally had a lower quality than the images in the test set. These were scaled up when resizing in the preprocessing step and therefore, the test set is inherently more difficult to segment than the training set.

## IV. EXPERIMENTAL DESIGN, MATERIALS AND METHODS

The study focused on obtaining high-quality scanned images of bank checks from existing public datasets online. The criteria for including an image in the dataset was that the image must show a complete bank check and contain at least one handwritten signature in the signature field on the check. The signature must also be legible with sufficient contrast from the background. Almost 70% of the images were then acquired from an existing online database of bank check images, published by a research department of the Reserve Bank of India. Rest of the images were obtained from other real-life scenarios including counterfeit checks used in a banking scam in Russia, and dummy checks filled by the author himself and captured using a mobile phone camera, or scanned with a scanner. The checks were obtained from a wide array of sources with different collection methods to ensure the variation in the quality, lighting condition and other parameters in the dataset to reflect the real world conditions.

The GIMP image processing software was used to manually label each signature pixel in the obtained images, to create the pixel-level segmentation mask. This process was carried out over a series of several weeks, painstakingly labelling a few images every day. For each image, an empty black-colored canvas of the same dimensions as the corresponding image was created and overlaid by the check image. Then, each pixel which corresponded to a signature on the check image was filled with white color. The segmentation mask was then saved as a black-and-white JPEG image.

To generate patch-level segmentation masks, a Python script using OpenCV was written to estimate bounding boxes around the signature pixels in the pixel-level masks. These bounding boxes were then filled with white color, and saved as a separate black-and-white segmentation mask at patch level. The following short Python script was used for this purpose.

```
import cv2
import numpy as np

def patch_mask(im):
    p = np.fliplr(np.argwhere(im==255))
    x,y,w,h = cv2.boundingRect(p)
    patch=np.array([[x,y],[x+w,y],
                    [x+w,y+h],
[x,y+h]],
                    dtype=np.int32)
    cv2.fillPoly(im, [patch], 255)
    return im
```

For the images which contained multiple and non-overlapping signatures, bounding boxes were drawn manually in the GIMP software.

## ETHICS STATEMENT

A major part of the dataset is made up from an existing, publicly available dataset [2] of bank checks with signatures obtained from nine volunteers, with permission to reuse. The rest of the dataset consists of either counterfeit or sample bank checks from the Internet, which is completely anonymized. Where signatures from any real individuals were used, an informed consent was obtained from each participant before collecting their data.

## ACKNOWLEDGMENT

I would like to thank the Institute of Development and Research in Banking Technology, established by the Reserve Bank of India, for making public their dataset of bank check images which were used in generating a major part of this dataset.

## REFERENCES

1. Muhammad Saif Ullah Khan, "BCSD: Bank Checks Segmentation Database." Kaggle, 2020, doi: 10.34740/KAGGLE/DS/1041694.
2. Dansena, P., Bag, S. and Pal, R., 2017, December. Differentiating pen inks in handwritten bank cheques using multi-layer perceptron. In International Conference on Pattern Recognition and Machine Intelligence (pp. 655-663). Springer, Cham.